\documentclass[runningheads]{llncs}
\usepackage[T1]{fontenc}

\usepackage{graphicx}
\usepackage{multicol}
\usepackage{multirow}
\usepackage{amsmath}
\usepackage{amssymb}
\usepackage{graphicx}
\usepackage{subcaption}
\usepackage{booktabs}
\usepackage{dirtytalk}

\usepackage{acro}
\usepackage{appendix}

\DeclareAcronym{ocr}{
  short=OCR,
  long=Optical Character Recognition,
}
\DeclareAcronym{gt}{
    short=GT,
    long=Ground Truth
}
\DeclareAcronym{iou}{
    short=IoU,
    long=Intersection over Union
}
\DeclareAcronym{nmt}{
    short=NMT,
    long=Neural Machine Translation
}
\DeclareAcronym{cer}{
    short=CER,
    long=Character Error Rate
}
\DeclareAcronym{ber}{
    short=BER,
    long=Box Error Rate
}
\DeclareAcronym{ece}{
    short=ECE,
    long=Expected Calibration Error
}

\begin{document}
\title{Confidence-Aware Document OCR Error Detection}
%
%
\author{Arthur Hemmer\inst{1,2}\orcidID{0009-0006-7550-2568} \and
Mickaël Coustaty\inst{2}\orcidID{0000-0002-0123-439X} \and
Nicola Bartolo\inst{1}\orcidID{0000-0001-8049-9611} \and
Jean-Marc Ogier\inst{2}\orcidID{0000-0002-5666-475X}}
\authorrunning{A. Hemmer et al.}
%
\institute{Shift Technology, Paris, France \and
\email{\{arthur.hemmer,nicola.bartolo\}@shift-technology.com} \\
L3i La Rochelle, La Rochelle, France \\
\email{\{jmogier,mcoustat\}@univ-lr.fr}}
\maketitle              
\begin{abstract}
Optical Character Recognition (OCR) continues to face accuracy challenges that impact subsequent applications. To address these errors, we explore the utility of OCR confidence scores for enhancing post-OCR error detection. Our study involves analyzing the correlation between confidence scores and error rates across different OCR systems. We develop ConfBERT, a BERT-based model that incorporates OCR confidence scores into token embeddings and offers an optional pre-training phase for noise adjustment. Our experimental results demonstrate that integrating OCR confidence scores can enhance error detection capabilities. This work underscores the importance of OCR confidence scores in improving detection accuracy and reveals substantial disparities in performance between commercial and open-source OCR technologies.

\end{abstract}
\section{Introduction}
\ac{ocr} on scanned documents has significantly advanced due to developments in deep learning and computer vision. Despite these advancements, \ac{ocr} errors persist and negatively impact performance on downstream NLP tasks, especially on low-quality documents such as historical documents \cite{adesam2019exploring,cuper2023unraveling,fleischhacker2024improving,gupta2015automatic,hill2019quantifying}, but also on information retrieval \cite{de2023evaluating}, event detection \cite{boros2022assessing}, named entity recognition \cite{hamdi2020assessing,hamdi2023depth}, topic modeling \cite{mutuvi2018evaluating} and others \cite{todorov2022assessment,van2020assessing}.

Some \ac{ocr} errors are \say{genuine} errors that even humans would struggle with based solely on the visual information. Most \ac{ocr} systems attempt to transcribe even illegible text, often resulting in gibberish. In contrast, humans utilize optical uncertainty and contextual cues to infer corrections or recognize text as unreadable. Recent advancements in \ac{ocr} technology incorporate more contextual information through language models, though challenges remain, particularly with numerically dense documents~\cite{hemmer2023estimating}, which comes from the lack of numeracy in language models \cite{spithourakis2018numeracy}. For industrial applications such as automated decision-making, prioritizing accuracy over coverage is crucial. It is better to hold off on a decision when there is too much uncertainty, than making decisions with lower accuracy. As such, additional information could be used to improve detection and possibly correction of illegible text, such that the overall system can better balance the accuracy-coverage trade-off.

While advancements in \ac{ocr} technology have been documented and made publicly available, effectively leveraging these improvements often remains confined to commercial entities with substantial computational resources and data. As we will show in Sec.~\ref{sect:data}, current open-source \ac{ocr} systems generally do not match the out-of-the-box performance of proprietary, commercial alternatives.

Historically, post-OCR processing methods have been employed to mitigate some limitations of \ac{ocr} systems \cite{jatowt2019post,ramirez2022post,rigaud2019icdar,topccu2022neural,yasin2023transformer}. These methods consist of an additional layer of post-processing to detect and correct errors in transcribed text. Although competitions have spurred research into post-OCR correction \cite{chiron2017icdar2017,rigaud2019icdar}, the datasets provided by these competitions typically consist only of OCR outputs aligned with corrected transcriptions, lacking other potentially informative features like OCR confidence scores. Some studies have indicated these scores are informative about the quality of transcription~\cite{cuper2023unraveling,gupta2015automatic,hill2019quantifying,nguyen2021survey}, though others have noted that deep-learning-based (OCR) methods often display overconfidence \cite{guo2017calibration}. In this work, we study in further detail the performance and confidence scores of several open-source and commercial \ac{ocr} systems and investigate methods for using these confidence scores for improving post-\ac{ocr} error detection.

Our contributions include:
\begin{enumerate}
    \item A method for aligning and comparing outputs from different \ac{ocr} solutions.
    \item An analysis of the confidence calibration error of several commercial and open-source \ac{ocr} solutions.
    \item A post-OCR error detection method that makes use of \ac{ocr} confidence scores to improve error detection.
\end{enumerate}


\section{Related Work}
The post-OCR processing framework typically adheres to the noisy channel model as introduced by Shannon \cite{shannon1948mathematical}. This model attempts to recover the original sequence or \say{word} $w$ from a noisy sequence of observed symbols $o$, with the goal to find $\hat{w}$ as

\begin{equation}\label{optestim}
    \hat{w}(o) = \arg\max_{w} p(w|o).
\end{equation}

Historically, directly estimating $p(w|o)$ is challenging due to limitations in data volume and training methods. Instead, the noisy channel model applies Bayesian inversion to decompose the probability into the likelihood and the prior, often referred to as the "error model":

\begin{equation}
    \hat{w}(o) = \arg\max_{w} p(o|w)p(w).
    \label{postprocessing}
\end{equation}


This decomposition forms the basis of many post-OCR processing strategies \cite{nguyen2021survey}. Following this, most post-OCR processing methods can be roughly divided into two categories: isolated-word and context-dependent.

Early post-OCR methods rely on isolated-word approaches that use dictionaries and error models to correct errors on a word-by-word basis \cite{brill2000improved,church1991probability}. These methods, however, struggle with \say{real-word} errors, which are incorrect words that still form valid dictionary entries (e.g., \say{post} mistaken for \say{cost}). Real-word errors account for approximately 59\% of OCR errors \cite{jatowt2019deep}.

To address these limitations, context-dependent approaches using language models and \ac{nmt} techniques have been developed \cite{amrhein2018supervised,hajiali2022generating,nguyen2020neural,ramirez2022post,topccu2022neural,yasin2023transformer}. These methods leverage contextual information to detect and correct also real-word errors. Notably, the 2019 ICDAR competition on post-OCR text correction highlights that the most effective methods combine BERT for error detection with \ac{nmt} for correction \cite{amrhein2018supervised,nguyen2020neural}, or use \ac{nmt} ensembles for both tasks \cite{ramirez2022post}. Other research explores a three-stage process involving candidate generation, weighting, and scoring, though it does not surpass the performance of the aforementioned strategies \cite{nguyen2018adaptive}.


Current state-of-the-art methods typically divide the problem into two phases: detection and correction \cite{rigaud2019icdar,nguyen2020neural}. Initially, one model identifies erroneous words in the OCR output, which are subsequently processed by a second model, often an \ac{nmt}-based solution, for correction.

Regarding the use of \ac{ocr} confidence scores in post-OCR processing, while some studies explore the use of these scores to assess overall transcription quality when no ground-truth data is available \cite{cuper2023unraveling,gupta2015automatic,springmann2016automatic}, their potential for detecting individual errors remains underexplored, likely because post-OCR datasets do not include this \ac{ocr} confidence data \cite{chiron2017icdar2017,rigaud2019icdar}. In this paper we describe the construction of such datasets and perform an evaluation of using several methods including a novel one which integrates the confidence scores with a language model.

\section{Data Acquisition} \label{sect:data}
This section outlines the data acquisition process employed to assess the utility of OCR confidence scores in post-OCR processing. We test a variety of OCR systems, both open-source and commercial, and align their outputs with the ground-truth transcriptions from multiple public datasets as well as one private dataset. The alignment of these outputs presents significant challenges due to the diverse formats and results produced by different OCR technologies, which will be detailed further. Additionally, we present statistics related to the datasets obtained through this process.

\subsection{Datasets}
We use three widely recognized public datasets for our analysis: CORD \cite{park2019cord}, FUNSD \cite{jaume2019funsd}, and SROIE \cite{huang2019sroie}. These datasets are frequently used for training and evaluating OCR systems \cite{arachchige2021unknown,kim2022ocr,olejniczak2022text,rotman2022detection,subramani2020survey} and consist of a diverse collection of scanned administrative documents, such as forms, invoices, and receipts. The majority of the text in these documents is printed, although some documents may include handwritten sections. Detailed statistics for each dataset are provided in Table \ref{tab:datasets}.

To complement these public resources, we include a private dataset consisting of various anonymized, scanned administrative documents. The inclusion of this private dataset helps us account for any biases that might arise if the public OCR systems were trained on the aforementioned public datasets.

\begin{table}[ht]
    \centering
    \begin{tabular}{lrrr}
    \hline
    \textbf{Dataset} & ~~\textbf{\#Docs} & ~~\textbf{Avg. \#Chars/Doc} & ~~\textbf{Avg. \#Boxes/Doc} \\
    \hline
    CORD & 1000 & 139 & 23 \\
    SROIE & 1000 & 676 & 117 \\
    FUNSD & 200 & 933 & 153 \\
    Private & 119 & 1430 & 262 \\
    \hline
    \end{tabular}
    \caption{Overview of datasets.}
    \label{tab:datasets}
\end{table}

As can be seen in Tab.~\ref{tab:datasets}, the datasets vary mostly in terms of number of characters per document. This is mostly due to the nature of the documents, as CORD, with 139 characters/doc on average, consists mostly of small receipts where text is often blurred out for reasons of anonymity or irrelevance to the task. SROIE contains similar receipts, but is more normalized (background removed and rotated correctly) and does not have any text blurred. FUNSD and the private dataset consist of larger A4-sized documents which, consequently, also contain more text. The additional context can be useful for better correction of OCR errors.

While the average number of characters per document change, we observe about the same ratio of character to box, which is around 6 for all datasets. This suggests a uniformity in the granularity of the annotated bounding boxes in the ground truth transcriptions. 

\subsection{OCR}
\label{sec:ocr}
Several \ac{ocr} systems are evaluated on the datasets. We choose a mix of open-source and commercial, cloud-based \ac{ocr}s for comparison: Microsoft Document Intelligence\footnote{https://learn.microsoft.com/en-us/azure/ai-services/document-intelligence/concept-read?view=doc-intel-4.0.0}, Amazon Webservices (AWS) Textract\footnote{https://aws.amazon.com/textract/}, Google OCR\footnote{https://cloud.google.com/use-cases/ocr?hl=en}, DocTR\footnote{https://github.com/mindee/doctr}~\cite{doctr2021}, EasyOCR\footnote{https://github.com/JaidedAI/EasyOCR} and PaddleOCR\footnote{https://github.com/PaddlePaddle/PaddleOCR} \cite{du2020pp}. The commercial ones are chosen as they are part of the largest and  most commonly used cloud platforms. The open-source \ac{ocr}s were picked based on ease-of-use (should provide end-to-end detection and recognition out of the box), popularity and reported performance. For a fair comparison, all \ac{ocr}s are used with their default, out-of-the-box settings.


In order to determine \ac{ocr} errors, the \ac{ocr} results need to be aligned with the \ac{gt} transcription. This is not trivial \cite{neudecker2021survey} as different \ac{ocr}s work at different levels of granularity such as word, line or paragraph-level, and have different strategies for special characters such backslashes, dashes and others. The naive approach would be to sort the bounding boxes vertically and horizontally for both the OCR output and the GT, but this produces noisy results because of the slight variation in the bounding box coordinates.

A more refined method that is commonly used in computer vision is to match boxes using the intersection-over-union (IoU) with a set threshold. However, this is not ideal for OCR bounding box alignment where, for example, the IoU of the boxes of \textit{"."} and \textit{"text."} would be low because of the low bounding box area of the period, making the intersection of the two relatively small although the period is part of the same box, which makes it hard to pick a good threshold. This problem is also recognized by \cite{baek2020cleval}, which proposes Pseudo-Character Centers (PCC) to solve the matching problem. PCCs take a bounding box and divide it up into $c$ equally spaced points according to the number of recognized characters in the box as an approximation to the real character centers. We found this approach to be too sensitive to the tightness of the bounding boxes, especially when the ground truth boxes are snapped closely around the words.

\begin{figure}
    \includegraphics[width=\linewidth]{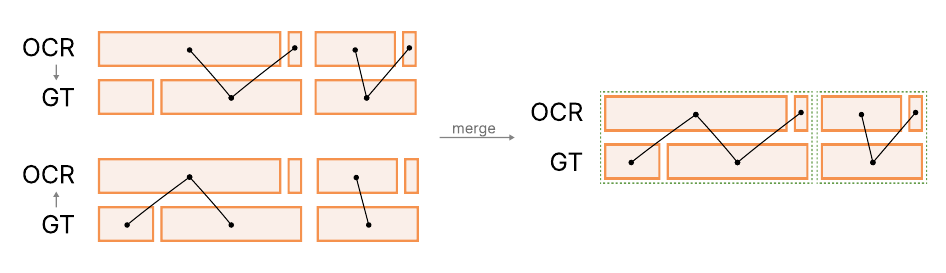}
    \caption{OCR and ground truth (GT) bounding box alignment strategy. First a corresponding GT box is found for each OCR box and vice-versa (left), the two mappings are merged and the connected components should contain the same information (right).}
    \label{fig:alignment}
\end{figure}

To solve the alignment problem, we propose a two-step solution as illustrated in Fig.~\ref{fig:alignment}. First, every predicted \ac{ocr} bounding box is matched to a \ac{gt} bounding box that covers at least 10\% of the OCR bounding box area and vice-versa. Doing it both ways ensures that the earlier illustrated case where \say{.} in \say{text.} is also taken into account. Although \say{.} may not cover 10\% of the \say{text.} bounding box, the bounding box of \say{text.} does cover more than 10\% of the bounding box of \say{.}. The 10\% was chosen empirically to filter out some matching noise, but we found that it could also be kept at 0 generally.

For the second step, the two mappings are merged into a single graph from which the connected components form the aligned boxes. The connected components are sorted according to the ground truth order. Unmatched \ac{gt} boxes are inserted in the final sequence. Unmatched \ac{ocr} boxes are left out however. While this means we will not capture \ac{ocr}s that predict too much, we found empirically that this is not often the case (rather the opposite), and when it is, it is often the ground truth that did not contain text from the background of a picture of a document. For completeness, we report the average number of unmatched \ac{ocr} boxes in our dataset statistics as well (see Tab. \ref{tab:ocrdatasets}).

Post-alignment, we determine the presence of errors in each OCR box by comparing its text with the GT text, converting all text to lowercase and removing spaces. When multiple OCR boxes are combined, their confidence score is calculated as the average of the individual box scores.

The overall \ac{cer} is computed by taking the Levenshtein distance between the \ac{ocr} and \ac{gt} sequences divided by the total number of \ac{gt} characters. In other words, it is the number of deletions $d$, insertions $i$ and substitutions $s$ divided by the total number of characters in the ground truth $n$:

\begin{equation}
    \text{CER} = \frac{d + i + s}{n}.
\end{equation}

\subsection{OCR Statistics}
\label{sect:stats}
Following the alignment and processing steps described above, we present various statistics for the different \ac{ocr}s and datasets in table \ref{tab:ocrdatasets}. In addition to the \ac{cer}, we also include the \ac{ber}, which is the percentage of connected components whose text is not equal to the \ac{gt} text. We furthermore compute the \ac{ece} \cite{guo2017calibration}, which is the average absolute deviation from the optimal calibration for each bin $B$:

\begin{equation}
    \text{ECE} = \sum_{i=1}^{m} \frac{|B_i|}{n} |y_i(B_i) - \hat{p}_i(B_i)|,
\end{equation}

Where each bin $B_i$ contains all predictions with confidence $(i-1)/10 \leq B_i < i/10$, $y_i(B_i)$ is the actual proportion of correct predictions in bin $B_i$ and $\hat{p}_i(B_i)$ the average confidence for the predictions in bin $B_i$. We set the bin size to $0.1$ for a total of $m = 10$ bins.

\begin{table}[ht]
    \centering
    \begin{tabular}{llrrrrr}
    \hline
     &  & ~~~\textbf{CER} & ~~~\textbf{BER} & ~~~\textbf{ECE} & ~~~\textbf{AC} & ~~~\textbf{AUB} \\
    \textbf{Dataset} & \textbf{OCR} &  &  &  &  &  \\
    \hline
    \multirow[t]{6}{*}{CORD} & AWS & 1.3 & 3.9 & 1.1 & 23 & 2.6 \\
     & Azure & 1.3 & 4.5 & 2.0 & 23 & 2.2 \\
     & Google & 5.3 & 9.9 & 2.3 & 23 & 3.4 \\
     & DocTR & 3.2 & 9.5 & 2.5 & 23 & 1.6 \\
     & EasyOCR & 17.4 & 50.4 & 10.3 & 18 & 1.6 \\
     & Paddle & 4.3 & 20.6 & 11.5 & 15 & 1.1 \\
    \hline
    \multirow[t]{6}{*}{SROIE} & AWS & 2.3 & 4.6 & 1.7 & 113 & 2.0 \\
     & Azure & 1.7 & 3.5 & 1.4 & 114 & 2.4 \\
     & Google & 2.8 & 5.8 & 3.0 & 115 & 3.4 \\
     & DocTR & 4.0 & 9.4 & 2.6 & 106 & 1.5 \\
     & EasyOCR & 12.0 & 44.2 & 15.5 & 69 & 0.5 \\
     & Paddle & 5.3 & 23.8 & 16.4 & 52 & 1.1 \\
    \hline
    \multirow[t]{6}{*}{FUNSD} & AWS & 3.1 & 6.3 & 3.0 & 161 & 4.6 \\
     & Azure & 2.3 & 5.9 & 3.0 & 162 & 6.1 \\
     & Google & 2.6 & 6.3 & 2.5 & 165 & 8.1 \\
     & DocTR & 7.7 & 19.7 & 5.4 & 151 & 4.0 \\
     & EasyOCR & 21.9 & 66.4 & 14.9 & 89 & 0.9 \\
     & Paddle & 8.9 & 36.5 & 22.6 & 54 & 0.6 \\
    \hline
    \multirow[t]{6}{*}{Private} & AWS & 1.4 & 2.0 & 2.1 & 214 & 6.2 \\
     & Azure & 0.8 & 1.5 & 2.5 & 216 & 4.2 \\
     & Google & 0.6 & 1.2 & 3.6 & 256 & 6.1 \\
     & DocTR & 5.4 & 15.4 & 3.7 & 205 & 2.4 \\
     & EasyOCR & 28.9 & 75.6 & 17.4 & 115 & 0.8 \\
     & Paddle & 4.5 & 19.7 & 11.7 & 51 & 0.4 \\
    \hline
    \end{tabular}
        
    \caption{OCR dataset statistics. CER = Character Error Rate, BER = Box Error Rate, ECE = Expected Calibration Error ($\text{bin size} = 0.1$), AC = Average number of Components, AUB = Average number of Unmatched OCR Boxes }
    \label{tab:ocrdatasets}
\end{table}

Table \ref{tab:ocrdatasets} highlights a significant performance gap between commercial and open-source OCR systems, with commercial OCRs generally outperforming their open-source counterparts. DocTR seems to be closest to the commercial OCRs, especially on the CORD dataset where it has a lower \ac{cer} than the Google OCR. Among the open-source OCRs, DocTR exhibits the lowest \ac{cer}, followed by PaddleOCR and EasyOCR. A qualitative analysis of OCR errors shows a frequent occurrence of errors on punctuation and special characters, especially among open-source ones. We hypothesize that the commercial OCRs might have specific post-processing steps for these types of errors, but we cannot verify this due to their black-box nature.

Performance on the private dataset mirrors that of the public datasets, with commercial OCRs achieving lower \ac{cer}. The similar or better performance of the OCRs on the private dataset compared to the public ones suggests that these OCR systems might not have been specifically trained or fine-tuned on the public dataset test sets, although we can not conclude this with certainty. We also note that the documents in the private dataset are of higher quality, which typically facilitates better OCR accuracy.

The \ac{ber} consistently exceeds the \ac{cer}, indicating that errors are typically isolated rather than clustered within the documents. The notably higher BER in EasyOCR and PaddleOCR across all datasets can be attributed to their tendency to generate larger, less precise bounding boxes, often encompassing lines or multiple words. This results from our alignment strategy, where more GT boxes are merged, increasing the likelihood of errors within a box, thereby raising the \ac{ber}.

Finally, we also observe that the proprietary \ac{ocr}s have a lower calibration error, although DocTR is showing similar calibration scores for CORD and SROIE as the commercial \ac{ocr}s. We expect the calibration error to correlate with the usefulness of integrating confidence score into an error detection model as we present in the next section.

\section{Confidence-Aware Error Detection}
\label{sect:method}
To integrate \ac{ocr} confidence scores into a post-OCR error detection model, we build on top of current state-of-the-art \ac{ocr} error detection work using BERT \cite{hajiali2022generating,nguyen2020neural} and make minimal modifications so that we would need to finetune the model as little as possible. BERT can be used for error detection by doing binary classification at the OCR box level, by labeling a box as erroneous if it contains an error.


\subsection{Architecture}
The confidence scores are integrated into the model by applying them directly to the initial token embeddings, much like positional encoding used in many transformer architectures. Given a list of tokens $\mathbf{t} = \{t_1, t_2, \ldots, t_n\}$ and an embedding function $ \textit{Emb}: t \longrightarrow \mathbb{R}^d $ where $d$ is the hidden dimension of the model, the confidence-aware embedding $e_i^c$ for token $t_i$ is obtained using

\begin{equation}
    e_i^c = (1 - \alpha)\text{Emb}(t_i) + \alpha (1 - p_{ocr}(t_i)).
\end{equation}

Where $p_{ocr}$ is the \ac{ocr} confidence in the box containing token $t_i$, broadcast across $\mathbb{R}^d$. The function $Emb$ here refers to the standard BERT embedding function which converts a token id (integer) into a vector in $\mathbb{R}^d$, before it is passed through the transformer part of the model.

The parameter $\alpha$ is a trainable parameter that controls how much the noise should be used in the model. As OCRs are calibrated differently (see Sec. \ref{sect:stats}), $\alpha$ was added to give the model more flexibility and potentially act as a knob to regulate the reliance on confidence scores or on the token embeddings. We investigate the impact of different values of $\alpha$ further in Sec.~\ref{sec:alpha}.



\subsection{Additional Pre-training}
In addition to integrating the confidence into the model, we also test an additional pre-training step to help the model learning how to integrate confidence scores. To do this, we continue pre-training a BERT model using the original Masked Language-Modeling (MLM) objective as well as a secondary binary noise prediction head. The pre-training uses data augmentation techniques where OCR noise is simulated on existing, non-noisy datasets.

The noise prediction head is used to predict whether a token was noised or not. We modify the MLM algorithm by sampling $p_{ocr} \sim \text{Beta}(4,1)$ and modify token $t_i$ for a random other token if $p_{ocr} < \text{Uniform}(0,1)$. The sampled $p_{ocr}$ is then used in the model to represent the confidence in the token. We choose to sample following a $\text{Beta}(4,1)$ as it skews the samples towards high probability which corresponds most to distribution of OCR confidence scores. For the final loss objective, the MLM and binary noise prediction cross-entropy losses are summed together.

We continue pre-training the BERT on the original datasets BookCorpus \cite{zhu2015aligning} and English Wikipedia for an additional 2.5k steps using the AdamW optimizer with a linearly declining learning rate of $5e-5$.

\subsection{Experiments}
We evaluate the proposed methods with the data acquired and aligned as described in section \ref{sect:data}. To compare with current state-of-the-art in error detection, we use an unmodified BERT model to classify boxes into \say{error} and \say{no error} as described in \ref{sect:method}.

The BERT and ConfBERT models are trained for a maximum of 16 epochs on the training split for each dataset using an AdamW optimizer with a learning rate of $5e-5$. Early stopping is used with a patience of 5 epochs and the checkpoint with the highest validation $F_1$-score is used for the final evaluation on the test set. Each training is repeated 10 times to measure the stability and significance of the obtained $F_1$-scores. We use the micro $F_1$-score and test statistical significance using the Kolmogorov-Smirnov test.

A baseline is calculated by relying solely on the \ac{ocr} confidences. It is computed by taking the percentiles of the confidences on the training set and then picking the threshold that results in the highest $F_1$ score on the validation set. The final score is evaluated on the test set for at the previously determined threshold. The results for the baseline and the other models can be found in Tab. \ref{tab:mainresults}.

Overall we observe that the error detection $F_1$ scores are higher for the open-source datasets, although this can mostly be attributed to the higher BER (see Tab. \ref{tab:ocrdatasets}) as this means lower class-imbalance and thus more positive-class training data, leading to an easier error detection overall. Due to these differences and the nature of the $F_1$ score, it is difficult to draw further conclusions from the $F_1$ scores between OCRs on the same dataset.

As to our initial hypothesis whether \ac{ocr}-confidence scores can be used to improve error detection, it seems that this is indeed the case. Among the results, we observe that integrating the confidence scores either increases or does not impact the $F_1$-score compared to the BERT base model. The exceptions being EasyOCR+Private and Paddle+FUNSD, where the ConfBERT models scores several points lower than the BERT base model. This can partially be explained by the high ECE computed for these combinations. Furthermore, the low granularity of these OCRs makes the confidence scores non-local, meaning that even if it had low confidence in a single character, this confidence might not be propagated to the predicted box spanning the whole line.


The improvements in $F_1$ score with ConfBERT are not always significant and the simple \ac{ocr}-confidence-only baseline can outperform a more complex method (AWS+CORD, Azure+Private, Google+Private, Paddle+Private, Paddle+FUNSD). The improvements over the BERT model seem to be highest for well-calibrated \ac{ocr}s. 


In terms of the additional pre-training we find that it mostly does not decrease the performance with respect to the non pre-trained models (except for Google+CORD), but it also does not improve it significantly in many cases. However, we recognize that there are other ways to integrate and pre-train for confidence-awareness which we have not explored, which might improve more. 

\begin{table}[ht]
    \centering
    \begin{tabular}{llllllll}
    \hline
     &  & ~AWS~ & Azure~ & Google~ & DocTR~ & EasyOCR~ & Paddle~ \\
    Dataset & Model &  &  &  &  &  &  \\
    \hline
    \multirow[t]{4}{*}{CORD} & Baseline & \textbf{0.45} & 0.44 & 0.34 & 0.47 & 0.73 & 0.47 \\
     & BERT & 0.23 & 0.47 & 0.46 & 0.53 & 0.86 & 0.69 \\
     & ConfBERT & 0.34* & \textbf{0.53*} & \textbf{0.56*} & 0.60* & \textbf{0.87*} & \textbf{0.70} \\
     & ConfBERT + Pretrain & 0.34* & \textbf{0.53*} & 0.51* & \textbf{0.62*} &\textbf{0.87*} & \textbf{0.70*} \\
    \hline
    \multirow[t]{4}{*}{SROIE} & Baseline & 0.39 & 0.42 & 0.30 & 0.46 & 0.70 & 0.51 \\
     & BERT & 0.38 & 0.31 & 0.41 & 0.67 & 0.86 & 0.74 \\
     & ConfBERT & 0.47* & \textbf{0.44*} & 0.51* & 0.70* & 0.87* & \textbf{0.75} \\
     & ConfBERT + Pretrain & \textbf{0.48*} & \textbf{0.44*} & \textbf{0.52*} & \textbf{0.71*} & \textbf{0.88*} & \textbf{0.75*} \\
    \hline
    \multirow[t]{4}{*}{FUNSD} & Baseline & 0.35 & 0.31 & 0.29 & 0.57 & 0.82 & \textbf{0.53} \\
     & BERT & 0.27 & 0.31 & 0.30 & 0.73 & 0.86 & \textbf{0.53} \\
     & ConfBERT & 0.36* & 0.36* & 0.32* & \textbf{0.75*} & 0.86 & 0.51* \\
     & ConfBERT + Pretrain & \textbf{0.37*} & \textbf{0.38*} & \textbf{0.34*} & \textbf{0.75*} & \textbf{0.88*} & 0.51* \\
    \hline
    \multirow[t]{4}{*}{Private} & Baseline & 0.31 & \textbf{0.35} & \textbf{0.22} & 0.52 & 0.88 & \textbf{0.47} \\
     & BERT & 0.22 & 0.17 & 0.07 & 0.83 & 0.89 & 0.30 \\
     & ConfBERT & 0.27* & 0.22 & 0.07 & 0.84* & 0.85* & 0.34 \\
     & ConfBERT + Pretrain & \textbf{0.36*} & 0.29* & 0.11 & \textbf{0.86*} & \textbf{0.92*} & 0.37* \\
    \hline
    \end{tabular}
    \caption{$F_1$ Error detection scores. \textbf{Bold} indicates the best score for the given dataset+OCR combination. * indicates statistically significant ($p < 0.05$) compared to BERT.}
    \label{tab:mainresults}
\end{table}

\subsection{Impact of $\alpha$} \label{sec:alpha}
\begin{figure}[!htb]
    \centering
    \includegraphics[width=\linewidth]{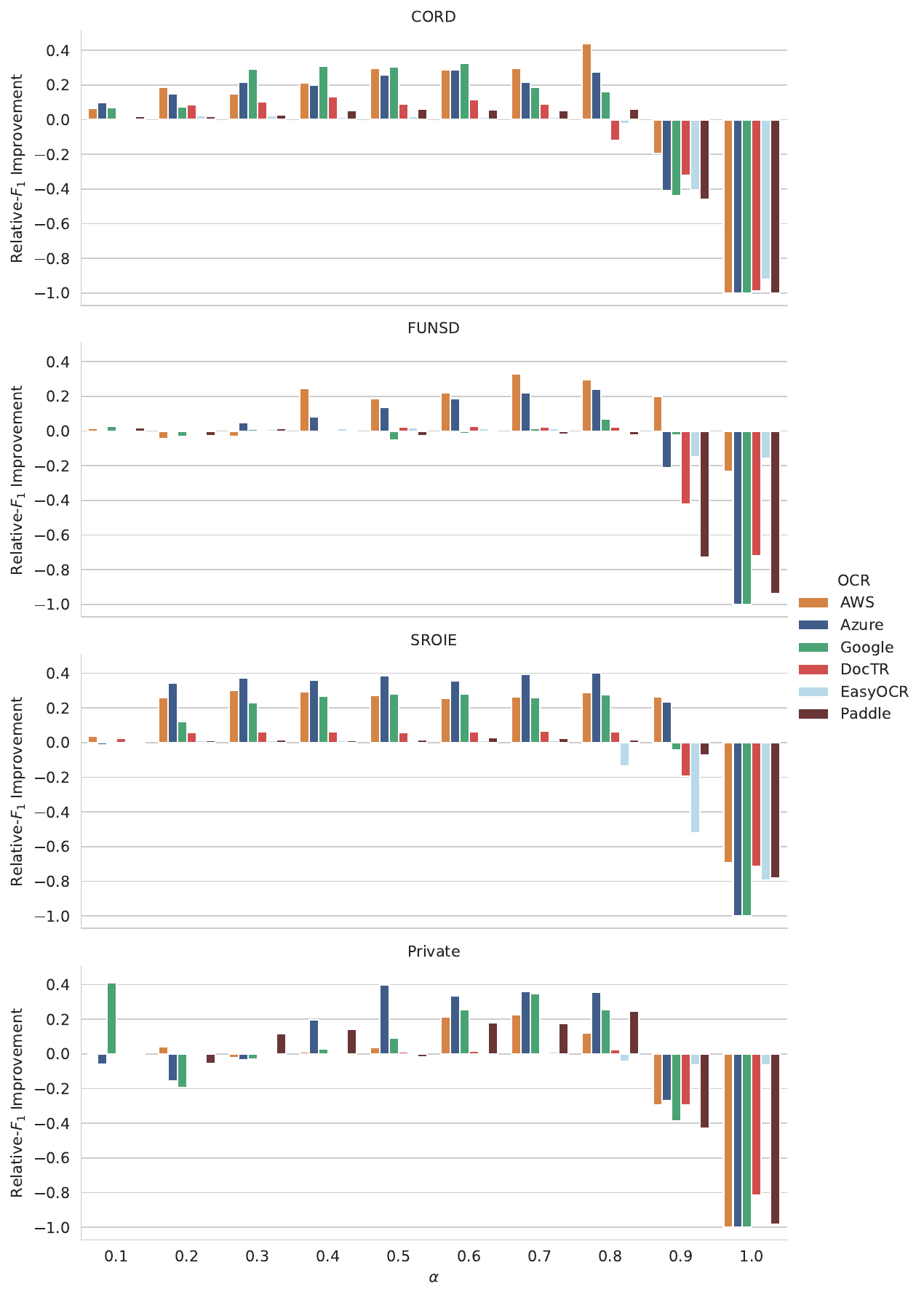}
    \caption{Relative improvement in $F_1$ for different values of $\alpha$ compared to $\alpha = 0$.}
    \label{fig:alpha}
\end{figure}

Although we set $\alpha$ to be a trainable parameter, the low learning rate and few training steps that the model goes through means the parameter can not change by much. As such, we investigate the impact of different values for $\alpha$ on the $F_1$ score. For this experiment, we fix $\alpha$ at a specific value in intervals of $0.1$ and make it non-trainable. The measured metric is the relative improvement of the $F_1$ score with respect to $\alpha = 0$. The results are shown in Fig. \ref{fig:alpha}.

As observed in the results of the main experiment, integrating confidence scores generally tend to improve or at least maintain the $F_1$ base scores. However, we note that starting from $\alpha = 0.9$, the model starts to be impacted negatively by the confidence scores as the information of the text itself \say{disappears}. 

Besides these larger values of $\alpha$, the relative improvement of the $F_1$ score (averaged over all \ac{ocr}s) is positively correlated with the value of $\alpha$, except for SROIE ($\text{p-value} = 0.09$ using Pearson correlation).

\section{Limitations}
While this work contains various metrics of several OCR systems, it is not intended to serve as an overall OCR benchmark. As detailed in Sec.~\ref{sec:ocr}, the OCR systems were chosen according to ease-of-use, popularity and reported performance. but many other OCR methods could have been considered and might have shown better results. The intention of this work was to study the informativeness of confidence scores among a variety of OCR systems.

Similarly, although we have shown that proprietary OCR systems achieve better \ac{cer}, an important benefit of open-source OCR systems is transparency and be able to adapt and tune a solution to specific needs. Simply changing the default parameters to better match the specific use-case can go a long way.

As for the confidence integration, we present a single way of integrating the confidence in the model. While we experimented briefly with some other ways, we found this to be performing the best. However, as the simple confidence-only baseline sometimes outperforms the other models, this suggests that there may be more effective ways to integrate this information into a model.

\section{Conclusion}
In this paper we investigated the possibility of using \ac{ocr} confidences for improving \ac{ocr} error detection. We created several datasets by running multiple commercial and open-source \ac{ocr}s on three public and one private dataset, and computed several metrics such as the error rate and calibration error by using a two-stage alignment algorithm. For multiple metrics, we notice an important gap between the commercial and open-source solutions which, however, can be partly attributed to the different sizes of bounding boxes the different OCR systems produce.

We build on top of current state-of-the-art error detection methods using BERT by adding the confidence scores to the embeddings. We find that using \ac{ocr} confidence scores mostly improves the results, although in some cases, for well-calibrated \ac{ocr}s, using the confidence scores with a simple baseline model might be enough.

This paper focuses on the task of error detection, but this is rarely done in isolation and is often paired with error correction. It would be interesting to investigate this in future work. Furthermore, there exist several methods for better calibrating model confidences. We find this an interesting avenue for further exploration.

\begin{credits}
\subsubsection{\ackname} This work was granted access to the HPC/AI resources of IDRIS under the allocation AD010614769 made by GENCI.

\end{credits}

%
%
%
\bibliography{biblio}
\bibliographystyle{splncs04}

\end{document}